\pdfoutput=1
\documentclass[11pt]{article}
\usepackage{acl}
\usepackage{times}
\usepackage{latexsym}
\usepackage[T1]{fontenc}
\usepackage{arydshln}
\usepackage{fixltx2e}
\usepackage[utf8]{inputenc}
\usepackage{enumitem}
\usepackage{microtype}
\usepackage{todonotes}
\usepackage{multirow}
\usepackage{color, colortbl}

\title{What's New? \\  Identifying the Unfolding of New Events in a Narrative}

\author{Seyed Mahed Mousavi\textsuperscript{*}, Shohei Tanaka\textsuperscript{\textdagger,\textdaggerdbl}, Gabriel Roccabruna\textsuperscript{*},\\
\textbf{Koichiro Yoshino\textsuperscript{\textdagger,\textdaggerdbl}}, \textbf{Satoshi Nakamura\textsuperscript{\textdaggerdbl}, \textbf{Giuseppe Riccardi\textsuperscript{*}}}\\
    \textsuperscript{\textdagger}Guardian Robot Project, RIKEN, Japan\\
    \textsuperscript{\textdaggerdbl}Nara Institute of Science and Technology, Japan \\
    \textsuperscript{*}Signals and Interactive Systems Lab, University of Trento, Italy\\
\texttt{mahed.mousavi@unitn.it,giuseppe.riccardi@unitn.it}}

\begin{document}
\maketitle
\begin{abstract}
Narratives include a rich source of events unfolding over time and context. Automatic understanding of these events provides a summarised comprehension of the narrative for further computation (such as reasoning). In this paper, we study the Information Status (IS) of the events and propose a novel challenging task: the automatic identification of \textit{new} events in a narrative. We define an event as a triplet of subject, predicate, and object. The event is categorized as new with respect to the discourse context and whether it can be inferred through commonsense reasoning. We annotated a publicly available corpus of narratives with the new events at sentence level using human annotators. We present the annotation protocol and study the quality of the annotation and the difficulty of the task. We publish the annotated dataset, annotation materials, and machine learning baseline models for the task of new event extraction for narrative understanding.
\end{abstract}

\section{Introduction}

The task of narrative understanding is a challenging topic of research and has been studied in numerous domains \cite{piper2021narrative, sang2022survey}. Recent studies include important applications of this task in supporting professionals in mental health. \cite{tammewar-etal-2020-annotation,adler2016incremental,danieli2022assessing}. Automatic narrative understanding may provide a summarized comprehension of the users' recollections that can be used to engage in personal and grounded dialogues with the narrator. Narrative understanding has been approached in different ways \cite{kronenfeld1978scripts,chambers-jurafsky-2008-unsupervised,kim-klinger-2018-feels}. A research direction in this field focuses on extracting the sequence of events that are mentioned in the narrative to obtain a summarized understanding of the whole narrative and its characters \cite{chen-etal-2021-event,mousavi-etal-2021-unsupervised}. In these works, the event is mostly represented by a predicate along with its corresponding subject and object dependencies. This definition relies on two assumptions a) the predicate represents an action/occurrence relation between the subject and the object dependencies; b) reoccurring characters across different events are the protagonists of the narrative. 

There have been interesting studies on different aspects of events in a narrative such as linking the correlated events as a chain \cite{chambers-jurafsky-2008-unsupervised}, learning semantic roles of participants \cite{chambers-jurafsky-2009-unsupervised}, common-sense inference~\cite{rashkin-etal-2018-event2mind}, and temporal common-sense reasoning \cite{zhou-etal-2019-going}. 

In order to obtain a concise and salient understanding of the narrative through the events, it is necessary to identify and select the events that relate to a new happening/participant in the narrative and have novel contributions. The process of recognizing a new event implicitly involves the event coreference resolution task, which consists of detecting the mentions of the same event throughout the content \cite{zeng-etal-2020-event}. Essentially, an event that is referring to a previous event is not considered new. Nevertheless, even if an event appears in the narrative for the first time it might be part of commonsense knowledge, and thus not provide any new information. 

In this paper, we address the problem of identifying new events as they unfold in the narrative. This task is inspired and motivated by the need to a) extract salient information in the narrative and position them with respect to the rest of the discourse events and relations, and b) acquire new events from a sequence of sentential units of narratives. This task can facilitate higher levels of computation and interaction such as reasoning, summarization, and human-machine dialogue. Last but not least, we believe this task is a novel and very challenging machine learning task to include in natural language understanding benchmarks. 

\begin{figure}[t]
	\begin{center}
	\includegraphics[width=7cm]{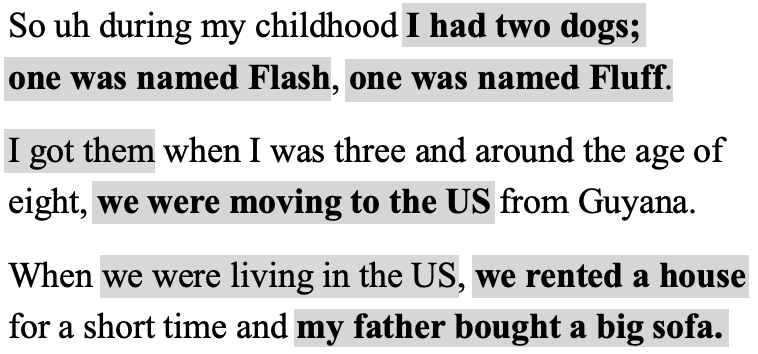}
	\end{center}
	\caption{An example of a narrative and the corresponding events. There are eight events in the sentences (highlighted), while six of them are presenting new information (bold) and the remaining two are referring to the already-mentioned events in the context (not bold).}
	\label{fig:event_example}
\end{figure}

We assess whether an event is new in a narrative according to their Information Status (IS) \cite{ Prince88thezpg, mann1992discourse}. IS refers to whether a piece of information, which can be represented as an entity or other linguistic forms, is new or old. We consider an event new if it has not been previously observed in the context and provides novel information to the reader; that is, its information (the event and/or participants) is not presented priorly in the discourse stretch, and it can not be inferred through commonsense. For instance, \textit{Bob \textbf{saw} Alice} is a new event if it is the first time that Alice is introduced in the narrative or the first time Bob saw her. However, once this event is selected as new, \textit{Bob \textbf{looked at} Alice} will not be a new event anymore. Furthermore, if \textit{Bob \textbf{married} Alice} is considered as a new event, \textit{Alice \textbf{is} Bob's wife} can be inferred through commonsense and thus is not a new event. An example of new and old events is presented in Figure~\ref{fig:event_example}. While there are eight events in the narrative sentences, two of them do not represent any novel information and thus are not new.

For this purpose, we developed an unsupervised model to extract markable event candidates from the narratives. We parsed a publicly available dataset of narratives, SEND~\cite{8913483}, and using the developed model, extracted all the markable events for each sentence. In the next step, we designed and conducted an annotation task using five human annotators to select the events in each sentence that are discourse-new with respect to the narrative context. In order to validate the annotation protocol and evaluate the results, we developed several neural and non-neural baselines for the task of new event extraction in both candidate-selection and sequence-tagging settings.

The contributions of this paper can be summarized as follows:
\begin{itemize} [noitemsep,topsep=2pt,parsep=2pt,partopsep=2pt]
    \item We present the novel task of new event detection for narrative understanding along with its annotation methodology and evaluation.
    \item We present the annotated version of a public corpus of emotional narratives for the task of automatic detection of new events in a narrative. 
    
    \footnote{\href{https://github.com/sislab-unitn/New-Event-Detection}{Link to our Repository}}.
    \item We introduce several baseline benchmarks for the task of new event detection based on discourse heuristics and deep neural networks, in two different settings of candidate selection and sequence tagging. 
\end{itemize}

\section{Literature Review}

\textbf{Event Extraction} The definition of the event concept has been the topic of study in different disciplines, originating in philosophy \cite{Mourelatos1978Events}. Early attempts to understand the semantics and structures of events in the text used hand-coded scripts with predefined slot frames to be filled by the values extracted from the text \cite{kronenfeld1978scripts}. This approach was later adopted by other works \cite{kim-klinger-2018-feels,ebner-etal-2020-multi}. \citet{kim-klinger-2018-feels} consider the activation of emotions as an event and study such events through different properties such as cause, experiencer, target, etc. In this definition, not only verb phrases but also noun phrases and prepositional phrases that manifest an emotion in a narrative participant can represent events. \cite{ebner-etal-2020-multi} studied the events and their participants by the verb-specific roles the participants can have (the arguments of the event "attack" are of types "attacker" and "target"). In this work, the authors formalized the event understanding as an argument-linking task. 

To address the expensive nature of designing domain-specific frames, \citet{chambers-jurafsky-2008-unsupervised} proposed an unsupervised approach to extract the event chains in a narrative according to the linguistic structures of the narrative sentences. Based on the assumption that reoccurring participants among different events are the protagonists of the narrative, the authors defined an event in a sentence as a predicate (verb) and the verb dependencies including the protagonist. This work was complemented further by considering the role of the protagonists in each event and the neighboring events in order to obtain a schema~\cite{chambers-jurafsky-2009-unsupervised}.

\textbf{Event-Centric Understanding} There have been several studies on the application of event-centric narrative understanding. \citet{mostafazadeh-etal-2016-corpus} studied the understanding of commonsense stories via event chain extraction model \cite{chambers-jurafsky-2008-unsupervised}. \citet{rashkin-etal-2018-event2mind} conducted a task on inferring the next possible intents and reactions of the participants in a narrative based on the observed events through commonsense. \citet{zhou-etal-2019-going} studied the application of temporal reasoning such as order/frequency of events in the narrative for the question-answering setting. \citet{mousavi-etal-2021-unsupervised} extracted events in a personal narrative to construct the personal space of events and participants in the user's life as a graph.

\textbf{Event Co-reference Resolution} The event coreference resolution task is focused on identifying the events that refer to previously mentioned events in a context. Two events are considered identical if they share the same spatiotemporal location \cite{quine1985events}. \citet{bejan-harabagiu-2010-unsupervised} studied the detection of coreferential events by measuring the similarity among two events using lexical and semantic features. \citet{zeng-etal-2020-event} proposed a model based on BERT pre-trained model \cite{devlin-etal-2019-bert} to integrate event-specific paraphrases and argument-aware semantic embeddings for this task.

\section{Definition of New Event}

We introduce the task of identifying the new events in a narrative to obtain a distilled and concise representation of the whole narrative and its characters. We follow the definition of an event that was used by \citet{chambers-jurafsky-2008-unsupervised} based on the verb and its dependencies. That is, a verb is a core element of an event and supports the relation among its dependencies such as subject, object/oblique nominals which are considered as the participants of the event~\cite{mousavi-etal-2021-unsupervised}. 

\citet{Prince88thezpg} defined the notion of old or new Information Status (IS) with respect to two aspects of the hearer's beliefs and the discourse model. New information according to the hearer's belief is the one that is assumed not to be already known for the hearer, while discourse-new information is the one that has not been mentioned or has not occurred priorly in the discourse-stretch \cite{Prince88thezpg}. \citet{nissim-etal-2004-annotation} adopts the IS concept and defines three categories of old, new, and mediated for the status of entities in a dialogue. The notion of old follows the definition provided by \citet{Prince88thezpg} closely. However, the authors define mediated as entities that have not been introduced directly in the context but are inferrable or generally known to the hearer; while the new category spans over entities that are not introduced priorly in the dialogue context, nor can they be inferred from the previously mentioned entities.

We extend the definition of the new category in entities \cite{nissim-etal-2004-annotation} to events. We define new events as those that are not mentioned in the narrative context and can not be inferred through commonsense by the reader. In this work, we do not consider further distinctions such as  old or mediated.

\section{Annotation of New Event}

\subsection{Annotation Task Description}

\textbf{Narrative Dataset} We conducted an annotation task for identifying the new events in narratives at the sentence level. The corpus used in this study is the SEND dataset \cite{8913483}, which is a collection of emotional narratives. The dataset consists of 193 narratives from 49 subjects, collected by asking each narrator to recount 3 most positive and 3 most negative experiences of her/his life. The statistics of the SEND dataset are presented in Table~\ref{table:sendstatistics} (the train, valid, and test sets are the official splits). 

\begin{table}
\centering
\small
\begin{tabular}{lc}
 &\textbf{Value}  \\
\hline\hline
\textbf{\#Narratives} (Train:Valid:Test)   & 193  (114:40:39)\\
\hline 
\textbf{\#Subject} (\# female)         & 49 (30)        \\
\hline
\textbf{Avg. Narrative Len.  }         & 28.10 utterances       \\
\hline       
\textbf{Avg. Utterance Len.  }         & 15.44 tokens       \\
\hline       
\textbf{\#Vocabulary}         & 4,416 unique tokens       \\
\hline
\end{tabular}
\normalsize
\caption{The statistics of SEND dataset \cite{8913483}. The dataset is provided with official train, valid and test sets. The majority of narrators are female and each narrative consists of approximately 430 tokens on average.}
\label{table:sendstatistics}
\end{table} 

\begin{figure*}[t]
	\centering
	\includegraphics[scale=0.24]{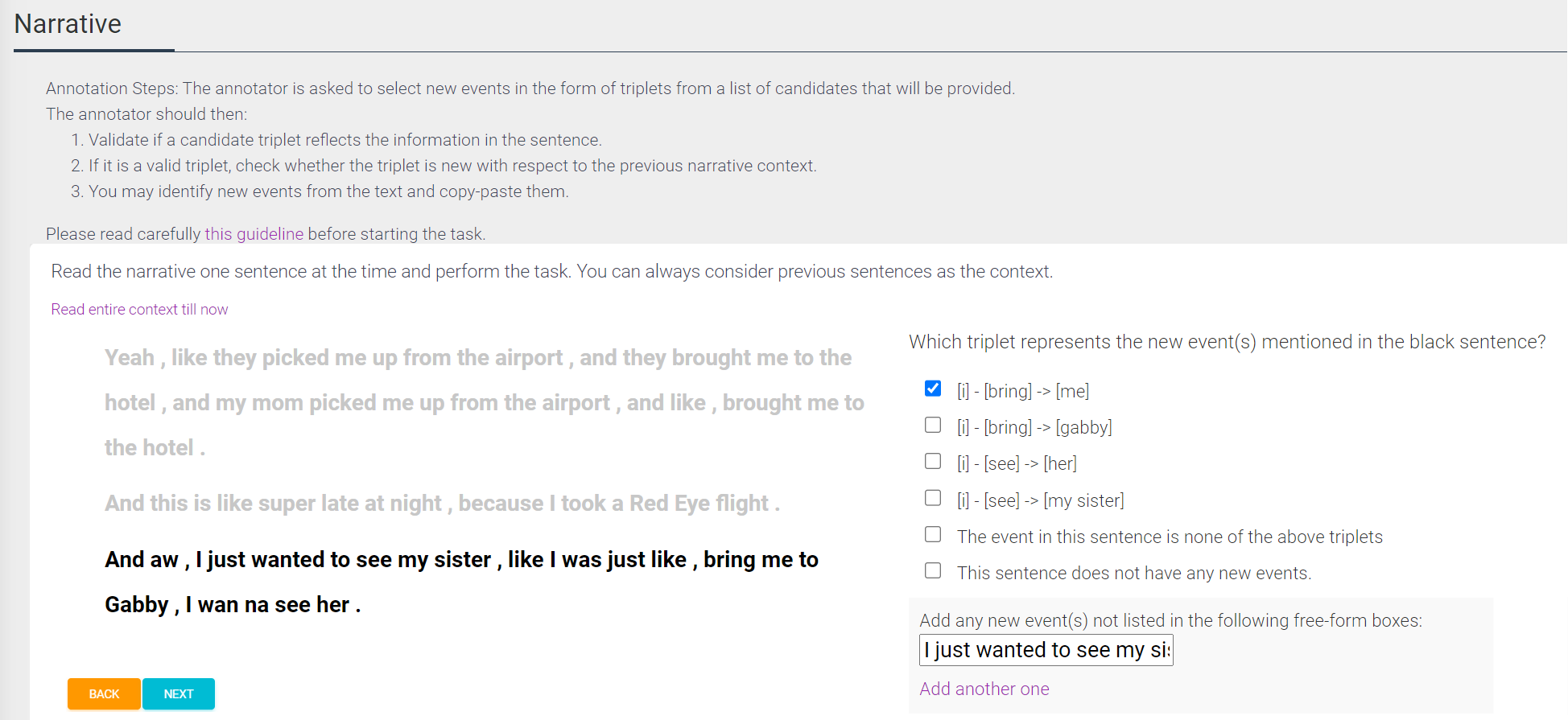}
	\caption{The user interface of the annotation platform. The annotator is presented with the narrative one sentence at a time on the left side of the screen. The event candidates and the option to add new events as free-from text are located on the right side of the interface. Moreover, a short version of the guidelines and the previous context of the narrative are shown to the annotator throughout the annotation.}
	\label{fig:platform_1}
\end{figure*}

\textbf{Task Design} To reduce the annotators' workload, we developed a baseline model inspired by \citet{mousavi-etal-2021-unsupervised} to automatically parse and extract all event candidates for each sentence in the narrative as the triplets of (subject, predicate, object). In the cases where more than 5 candidates were extracted for a sentence, we created 5 clusters using Levenshtein distance \cite{yujian2007normalized} (hierarchical clustering) and the candidate with the most number of tokens in each cluster was selected to be presented to the annotator. We randomly sampled 21 narratives from the SEND dataset and reserved them as backup data (13 narratives from the train set, 4 from the valid set, and 4 from the test set). Using the extraction pipeline, we extracted all subject-predicate-object triplets as event candidates in the remaining 172 narratives at the sentence level.

\textbf{Annotation UI} The user interface (UI) of the annotation platform is presented in Figure \ref{fig:platform_1}. Throughout the task, the annotator is presented with a brief version of the task guidelines on the top of the display (with access to the complete version). The narrative is presented on the left side of the screen with the current sentence in black and the context in grey. The narrative is updated progressively sentence-by-sentence while the annotator has access to the previous sentences of the context. For each sentence, the annotation question, the list of the triplet candidates and the possibility to select and add continuous span from the text are presented on the right side.

\textbf{Annotation Task} During the task, the annotators were presented with a narrative one sentence at a time and the corresponding list of candidates. They were asked to control if any of the candidate triplets in the list is valid (i.e. it reflects the information in the sentence correctly); and whether it provides new information with respect to the previous narrative context, that can not be inferred through commonsense. In the case of valid and new information, the annotators were asked to select that candidate as a new event. Furthermore, if there were no candidates extracted for a sentence or the new information in a sentence was not presented as a valid candidate, the annotator was asked to add the new information by simply copying the segment that conveys it from the sentence and adding it as continuous span text.

\textbf{Task Execution} We recruited five annotators for the task of new event annotation. The annotators were non-native English speakers with certified English proficiency. After an introductory meeting with the annotators, they were asked to carry out the first qualification task which consisted of annotating one narrative, sampled from the valid set. The result of the first qualification batch was checked manually and a few refinements were made with the annotators. The annotators were then asked to perform a second qualification task using another narrative randomly sampled from the valid set. The Inter-Annotator Agreement (IAA) level during the two qualification tasks, which is presented in Table~\ref{tab:qualification}, indicates the improvement in the annotators' performance from one qualification batch to the other. The IAA for the event candidates is calculated using Krippendoff's $\alpha$ \cite{krippendorff2011computing}, while the IAA for the continuous span text is calculated by the extension of Cohen's $\kappa$ for segmentation agreement \cite{fournier2012segmentation}, averaged among all annotators. The remaining 170 narratives were divided into 11 batches. In each batch, one narrative was annotated by all annotators for the purpose of continuous quality control of the results, while the rest was equally divided among the annotators. To prevent unreliable and biased agreements, all 11 overlapping narratives were from different narrators.

\begin{table}
\centering
\small
\begin{tabular}{lcc|c}
\textbf{} &\multicolumn{2}{c}{\textbf{Qualifications}} & \multirow{2}{*}{\textbf{Overall IAA }}\\
\textbf{Annotation Format} &\textbf{First} & \textbf{Second} & \textbf{ } \\
\hline\hline
{Selected Candidates}   & 0.22 & 0.55 & 0.54  \\
\hline 
{Added Spans}         & 0.32 & 0.60  & 0.66    \\
 \hline
\end{tabular}
\normalsize
\caption{Inter-Annotator Agreement (IAA) during the qualification tasks and over the whole annotation task. The results indicate an improvement in the performance of annotators from one qualification batch to the other. The IAA is computed for candidate selection and continuous span selection annotation using Krippendoff's $\alpha$ and the extension of Cohen's $\kappa$ for segmentation agreement, respectively.}
\label{tab:qualification}
\end{table} 

\subsection{Annotation Result Evaluation }

\begin{figure}[t]
	\begin{center}
	\includegraphics[width=7.5cm]{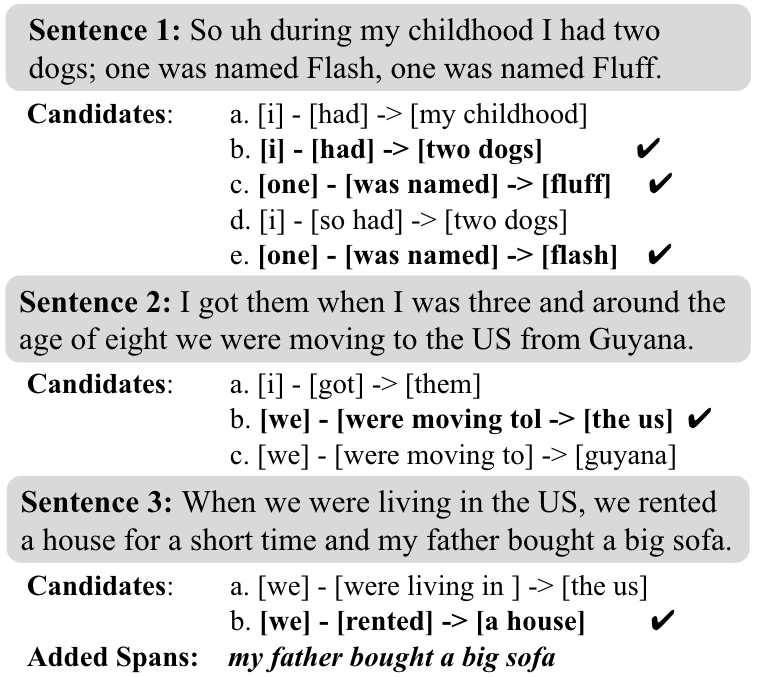}
	\end{center}
	\caption{An example of sentences in a narrative and the corresponding events; while the baseline model has extracted various event candidates, only a few of them are valid and new events (bold). Furthermore, the baseline model has missed an event in the third sentence which is added as a span from the sentence.}
	\label{fig:sentence_example}
\end{figure}

\begin{table}[t]
\centering
\begin{tabular}{lc}
\hline\hline
\multicolumn{2}{c}{\textbf{Selected New Events as Candidates}}\\
\hline\hline
 {\#Candidates selected}  & 1536 \\
 \hline
  {Avg. candidates selected:}&\\
  \hspace{1cm}{\textit{per Sentence}} & 0.57 \\
 \hspace{1cm}{\textit{per Narrative}} & 9.0 \\
  \hspace{1cm}{\textit{per Narrator}} & 31.4 \\
\hline 
{\%Candidates selected in:} & \\
  \hspace{0.5cm}{\textit{ 1\textsuperscript{st} half of the Sentence}} & 43\% \\
  \hspace{0.5cm}{\textit{ 2\textsuperscript{nd} half of the Sentence}} & 57\% \\
  \hspace{0.5cm}{\textit{ 1\textsuperscript{st} half of the Narrative}} & 55\% \\
  \hspace{0.5cm}{\textit{ 2\textsuperscript{nd} half of the Narrative}} & 45\% \\
\hline\hline
\multicolumn{2}{c}{\textbf{Added New Events as Continuous Spans}}\\
\hline\hline
 {\#Spans added}  & 2254\\
\hline 
{Avg. spans added:} & \\
\hspace{1cm}{\textit{per Sentence}} & 0.8 \\
 \hspace{1cm}{\textit{per Narrative}} & 13.3\\
  \hspace{1cm}{\textit{per Narrator}} & 46.0 \\
  \hline 
  {\%Spans added in:} & \\
  \hspace{0.5cm}{\textit{ 1\textsuperscript{st} half of the Sentence}} & 38.1\% \\
  \hspace{0.5cm}{\textit{ 2\textsuperscript{nd} half of the Sentence}} & 61.9\% \\
  \hspace{0.5cm}{\textit{ 1\textsuperscript{st} half of the Narrative}} & 96.9\% \\
  \hspace{0.5cm}{\textit{ 2\textsuperscript{nd} half of the Narrative}} & 3.1\% \\
\hline
\end{tabular}
\caption{The statistics of the annotated dataset. While only 1536 extracted candidates (out of 6938, thus 22\%) were selected as new events, 2254 new events were added by the annotators as continuous span text. Moreover, almost all of the continuous span events appear in the first half of the narrative, while event candidates have a quite normal distribution.}
\label{tab:annotation_statistics}
\end{table} 

\begin{figure*}[t]
    \centering
    \includegraphics[scale=0.37]{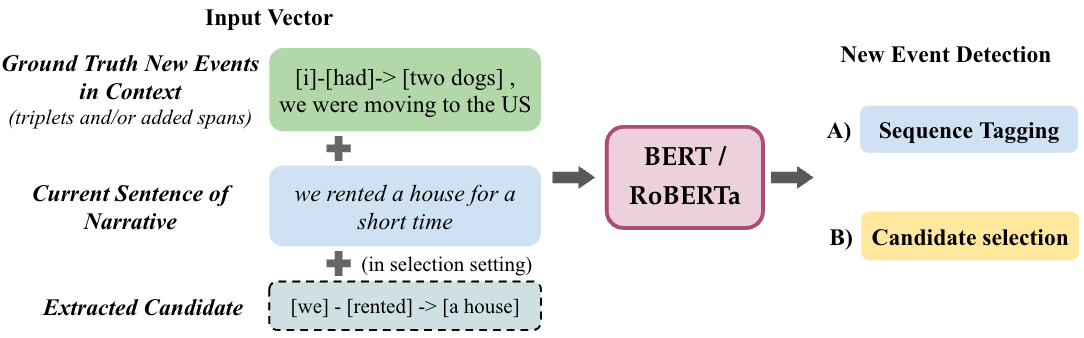}
    \caption{The neural baselines for the task of new event detection. The input vector consists of the new events in the context (ground truth) and the current sentence. In the candidate selection setting, the input vector includes the extracted candidate as an additional segment as well. The model encodes the input vector and outputs either a) a sequence of tags, corresponding to the tokens in the sentence; or b) a binary decision to categorize the candidate as new or not.}
    \label{fig:nn}
\end{figure*}

We annotated the dataset of personal narratives, SEND \cite{8913483}, with new events in the sentence level by five human judges. An example of the annotation results is presented in Figure~\ref{fig:sentence_example}. While the baseline model has extracted various possible event candidates from the sentence, only a few of them are \textbf{valid} events that are representing \textit{\textbf{new}} information. Moreover, the model has failed to extract an event in the third sentence which is added as a span from the text.

Throughout the task, the IAA level on the overlapping narratives was computed to ensure a consistent annotation quality. We observed negligible fluctuations in the IAA level during the task (<0.9 for Krippendoff's $\alpha$), except for one batch; for which the low-quality contributions were detected and refinements were made with one annotator. The overall IAA level of the annotated dataset is presented in Table \ref{tab:qualification}. The results are close to the level obtained in the second qualification batch.

The statistics of the annotated dataset, presented in Table \ref{tab:annotation_statistics}, indicate that the majority of the annotated events were added as continuous span text and were not extracted by the baseline model. Moreover, while the event candidates appear in the narrative with an approximately uniform distribution, almost all of the continuous span events are located in the first half of the narrative. This result is in line with the definition of new events since the events mentioned before in the context are "old" events. Nevertheless, in both cases of candidate events and continuous span events, we observe that the second halves of the sentences contain more information than the other half, indicating that the narrators tend to mention the new events at the end of the sentence.

\begin{table}[t]
\centering
\begin{tabular}{lccc}
\textbf{} &\textbf{Prec.} & \textbf{Rec.} & \textbf{F1} \\
\hline\hline
\textbf{Random} & 24.0 & 29.2 & 26.3 \\
\textbf{Binary} & 22.8 & 49.4 & 31.2 \\
\textbf{First Candidate} & 27.7 & 33.7 & 30.4 \\
\textbf{Last Candidate} & 30.1 & 36.7 & 33.1 \\
\textbf{New Subject} & 24.6 & 28.6 & 26.5 \\
\textbf{New Entity} & 25.1 & 88.9 & 39.1 \\
\hline
\textbf{BERT} & $35.6$ & $51.1$ & $41.6$ \\
\textbf{RoBERTa} & $40.4$ & $83.1$ & $\textbf{54.3}$ \\
\hline 
\end{tabular}
\normalsize
\caption{The results of the new event candidate selection baselines. The performance of the neural models is averaged over 10 runs.}
\label{tab:selection_results}
\end{table} 
 
\section{Baselines for New Event Detection}

We developed neural and non-neural baselines to validate the outcome of the annotation task, and, as baselines for the novel task of new event detection in a narrative. Considering the two annotation formats of selecting candidates and adding continuous spans, we formalize the task using two settings of candidate selection and sequence tagging.

\subsection{Candidate Selection Baselines}
The first group of models is tasked to select the new events from the candidates extracted by our baseline model. The rule-based models are:
\begin{itemize} 
        \item \textbf{Random Selector}: for each sentence and its event candidates, it randomly picks one candidate as the new event in the sentence.
        \item \textbf{Binary Selector}: for each of the event candidates of a sentence, it randomly decides whether it is a new event or not. Thus, each candidate has a 50\% chance of being selected as a new event. 
        \item \textbf{First Candidate Selector}: that selects the first event candidate that is extracted for a sentence as the new event. 
        \item \textbf{Last Candidate Selector}: which selects the last event candidate that is extracted for a sentence as the new event for the sentence.
        \item \textbf{New Subject Selector}: which selects the first candidate that contains a new (unseen) subject in the list of candidates as the new event. In other words, the number of selected candidates is equal to the number of non-repetitive subjects in the candidate list of the narrative.
        \item \textbf{New Entity Selector}: which selects all the event candidates that include new subjects or new objects at the narrative level. Thus, it selects all candidates unless they differ in the verb only. In that case, it selects one of them as the new event.
\end{itemize}

\textbf{Neural Network Models} In addition to the rule-based models, we developed neural models based on Pre-trained Language Models (PLMs) as baselines for the task of new event candidate selection presented in Figure \ref{fig:nn}. For this purpose, we model the input vector with three elements as event candidate, current sentence, and context new events. The context new events denote the new events (ground truth) in the narrative context up to the current sentence. In cases where the size of the input vector exceeds the model limits (for instance 512 tokens per BERT-based models), the model trims the former part of the context new events. The model encodes this vector and outputs the classification decision of whether the event candidate (triplet) is a new event or not. The PLMs we fine-tuned for this purpose are BERT \cite{devlin-etal-2019-bert}, and RoBERTa \cite{liu2019roberta}.

The results of the candidate selection baselines are presented in Table \ref{tab:selection_results}. We observe that \textit{Last Candidate Selector} has achieved the highest precision level among rule-based models. This is in line with the annotation result analysis, indicating the percentage of selected new event candidates to be slightly higher at the end of sentences. On the other hand, \textit{New Entity Selector} achieves the highest level of recall while having a very low level of precision, as it selects all candidates unless the variation is only in the verb predicate. Moreover, the F1 scores of all the rule-based models are less than 40.0\%. This indicates that features such as the novelty in elements or occurrence position are not enough to achieve high performance on the task of new event selection. While both neural models outperform the rule-based ones, RoBERTa outperforms all the baselines in this task by having the highest level of precision while maintaining a high recall.

\subsection{Sequence Tagging Baselines}

\begin{table}[t]
\centering
\begin{tabular}{lccc}
\textbf{} &\textbf{Prec.} & \textbf{Rec.} & \textbf{F1} \\ 
\textbf{} &\textbf{(\%)} & \textbf{(\%)} & \textbf{(\%)} \\ 
\hline\hline
\textbf{Random} & 18.8 & 49.7 & 27.3\\ 
\textbf{Early} & 17.4 & 29.5 & 21.9\\ 
\textbf{Late} & 20.2 & 34.0 & 25.4\\ 
\hline
\textbf{BERT} & 33.2 & 82.2 & 47.3 \\ 
\textbf{RoBERTa} & 34.3 & 81.3 & $\textbf{48.3}$ \\ 
\hline 
\end{tabular}
\normalsize
\caption{The results of the new event sequence tagging baselines. The models are trained and tested on continuous span events annotated by the human judges only. The performance of the neural models is averaged over 10 runs.}
\label{tab:tagging_results_pasted}
\end{table}


The second group of the models is developed for the task of new event detection in a sequence tagging setting. That is, the models tag the sequence of tokens (chunks) which are representing a new event in the sentence. The analysis performed on the continuous span events selected by the human judges indicated that several events can share the same tag spans such as subject or object. Therefore, we formalize this task as a binary tagging task rather than IOB tagging task and leave the development of the models for IOB tagging of multiple spans with overlap as future work. Similar to the previous task, we developed rule-based and neural baselines for new event sequence tagging. The developed rule-based baselines are:

\begin{itemize} 
        \item \textbf{Random Tagger}: which randomly tags tokens in a sentence as the new event tokens. 
        \item \textbf{Early Tagger}: which tags the tokens in the first 30\% of a sentence as the new event tokens. 
        \item \textbf{Late Tagger}: which tags the tokens in the last 30\% of a sentence as the new event tokens. 
\end{itemize}

\begin{table}[t]
\centering
\begin{tabular}{lccc}
\textbf{} &\textbf{Prec.} & \textbf{Rec.} & \textbf{F1}\\ 
\textbf{} &\textbf{(\%)} & \textbf{(\%)} & \textbf{(\%)} \\ 
\hline\hline
\textbf{Random} & 31.1 & 49.6 & 38.2\\ 
\textbf{Early} & 30.8 & 31.6 & 31.2 \\ 
\textbf{Late} & 29.9 & 30.4 & 30.2\\ 
\hline
\textbf{BERT} & 54.9 & 84.3 & 66.5\\ 
\textbf{RoBERTa} & 55.5 & 84.8 & \textbf{67.1} \\ 
\hline 
\end{tabular}
\normalsize
\caption{The results of the new event sequence tagging baselines. Compared to Table~\ref{tab:tagging_results_pasted}, in this setting, the models are trained and tested on both selected candidates and continuous span events annotated by the human judges. The performance of the neural models is averaged over 10 runs. }
\label{tab:tagging_results_triplet_pasted}
\end{table}

\textbf{Neural Network Models} Using BERT \cite{devlin-etal-2019-bert}, and RoBERTa \cite{liu2019roberta} PLMs, we developed two neural baselines for this task. The models take as input the current sentence and the context new events which are the sequences of new events in the narrative context up to the current sentence. Similarly to the previous neural baselines, if the input vector exceeds the size limits of the models the former part of the context new events is trimmed. The model encodes this vector and outputs a tag sequence consisting of \textbf{E}\textsubscript{(vent)} or \textbf{O}, corresponding to the tokens in the sentence, indicating whether or not they describe a new event.

We initially trained the sequence tagging baselines using the annotated continuous span events. The results of this experiment are presented in Table \ref{tab:tagging_results_pasted}. We observed that precision scores and consequently F1 scores are not significantly different among rule-based models. This indicates that the position of the tokens in the sentence is not the most contributing factor to the prediction accuracy. Similar to the previous task, the neural models have the highest performance among the baselines. However, their precision is considerably lower than the recall.

Similar to the previous task, the neural models have the highest performance among the baselines. However, their performance can be further improved by increasing the precision since it is considerably lower than the recall. The agreement level of the rule-based models is significantly small since the metric takes into consideration the beginning and the end of the tag spans. This is in contrast with the precision and recall metrics which focus on only binary values of each tag.

In the next step, we evaluated the same baseline models using both the selected event candidates and the continuous span annotations as the train and test sets. The results of this experiment, presented in Table \ref{tab:tagging_results_triplet_pasted}, show a boost in the performance of all models using the mentioned train and test sets. Nevertheless, the same performance trends among models can be observed in this experiment as well.

\section{Conclusions}

In this work, we study the events in narratives according to their Information Status. We introduce the new task of identifying new events as they unfold in the narrative. In our definition of the event, the verb is the central element that represents a relation/happening that engages its dependencies such as subject, object, or oblique nominals. Meanwhile, we define an event as new if it provides novel information to the reader with respect to the discourse (discourse-new) and if such information can not be inferred through commonsense. We annotated a complete dataset of personal narratives with new events at the sentence level using human annotators. We then developed several neural and non-neural baselines for the task of new event detection in both settings of candidate selection and sequence tagging. We share the annotated dataset and the baselines with the community. We believe this task can be a novel and challenging task in narrative understanding and can facilitate and support other tasks in natural language understanding, human-machine dialogue, and natural language generation.

\section{Limitations}
The dataset used in this work is a personal narrative corpus in English collected in-vitro (e.g. subjects in a lab setting). Further work will be needed to extend it to other languages, genres, and naturalistic conditions. The reproducibility of the annotation task may be subject to variability due to the fact that the task is done by five internal annotators and not through crowd-sourcing techniques.

\section*{Acknowledgements}
The authors would like to thank Desmond C. Ong for providing the dataset and useful discussions about the annotation task. 

\noindent We acknowledge the support of the MUR PNRR project FAIR - Future AI Research (PE00000013) funded by the NextGenerationEU.

\bibliography{custom}



\end{document}